# Hybrid Machine Learning Models for Crop Yield Prediction

Saeed Nosratabadi
*School of the built Environment*
*Oxford Brookes University*
Oxford, UK
0000-0002-0440-6564

Karoly Szell
*Alba Regia Technical Faculty*
*Obuda University*
Budapest, Hungary
0000-0001-7499-5643

Bertalan Beszedes
*Alba Regia Technical Faculty*
*Obuda University*
Budapest, Hungary
beszedes.bertalan@amk.uniobuda.hu

Felde Imre
*Kalman Kando Faculty of*
*Electrical Engineering*
*Obuda University*
Budapest, Hungary
0000-0003-4126-2480

Sina Ardabili
*Institute of advanced studies Koszeg*
*University of Pannonia*
Koszeg, Hungary
0000-0002-7744-7906

Amir Mosavi [1,2*]
[1] *Department of Mathematics and Informatics, J. Selye University*
Komarno, Slovakia
[2] *Bauhaus Universität Weimar,*
Weimar, Germany
0000-0003-4842-0613

*Abstract*—Prediction of crops yield is essential for food security policymaking, planning, and trade. The objective of the current study is to propose novel crop yield prediction models based on hybrid machine learning methods. In this study the performance of artificial neural networks-imperialist competitive algorithm (ANN-ICA) and artificial neural networks-gray wolf optimizer (ANN-GWO) models for the crop yield prediction are evaluated. According to the results, ANN-GWO, with R of 0.48, RMSE of 3.19, and MEA of 26.65, proved a better performance in the crop yield prediction compared to the ANN-ICA model. The results can be used by either practitioners, researchers or policymakers for food security.

## I. Introduction

Managing food security has turned into a very complicated and vital issue in the food supply chain [1]. Prediction of crop yield illustrates valuable possibilities for the management of food security in a food supply chain. Crop yield prediction presents the information that can be the basis of many important decisions related to food security, such as trading and developing policies [2].

On the other hand, forecasting the yield is not very easy as many controllable factors (e.g., applied irrigations, pest and fertilizer applications, etc.) and uncontrollable factors (e.g., weather, subsidies, and market, etc.) affect the crop yield [3, 4]. Numerous approaches were developed to predict the crop yield constituting farmers' long-term experience and the average of several previous yields. Whilst, Schlenker and Roberts [5] believe that the behavior of crop yield is not linear, and it varies from one year to another. The literature introduces data-driven models for crop yield production as the most accurate and efficient methods [6]. Although applying data-driven models for the crop yield prediction make accurate the data collection methods mechanisms, they are inexpensive and relatively easy to apply [6].

Therefore, the current study aims to find the most proper machine learning for crop yield prediction. To do so, a comparison study conducted to evaluate the performance of two hybrid machine learning methods artificial neural network imperialist competitive algorithm and artificial neural network gray wolf optimizer. This study took place in a large irrigated area in Kerman, Iran. In the following sections, the methodology and data collection process are described firstly. Then the performance of the mentioned machine learning methods is evaluated, and results and discussions consequently are provided.

## II. Material and Methods

### A. Case Study

The current study focused on the farms located near the city of "Kerman" in Iran. Kerman is the largest province in Iran (area of 183,285 km2), that embraces 11% of the land area of Iran (see Fig. 1). To meet the objective of the study, two types of data were collected: 1) agricultural production and 2) weather information. Spriter-GIS system used for the collection of data such as crop species, irrigation, and crop yield.

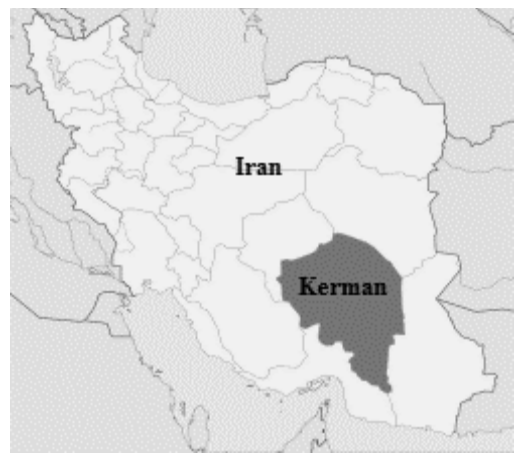

Fig. 1. Study area: Kerman, Iran

For the collection of data such as rainfall, solar radiation, and temperatures two meteorological stations placed on the site. On the other hand, agricultural products such as wheat, barley, potato, and sugar beets, which are the main productions of this region, were selected in this study. A summary of the collected data is presented in table I. As it is illustrated in table I, data recorded from 1998-2006 taken for further analysis in this study. The data related to 1998-2005 applied for the training phase and data 2005-2006 utilized for testing the methods' performance.

TABLE I. DISTRIBUTION OF SAMPLES DATASET IN TESTING AND TRAINING STAGES

| Crop species | Total number of samples | Samples in the training period (1999-2004) | Samples in the testing period (2005-2006) | Testing percentage (%) |
|---|---|---|---|---|
| Wheat | 508 | 449 | 59 | 11.61 |
| Barley | 87 | 42 | 45 | 51.72 |
| Potato | 195 | 132 | 63 | 32.31 |
| Sugar Beet | 156 | 108 | 48 | 30.77 |
| Total | 946 | 731 | 215 | |

It is worth mentioning that the attributes considered in the current study were planting area (ha), irrigation water depth (mm) (which refers to total consumed water volume during the six stages of crop growth), rainfall during the crop growth stages (mm), global solar radiation (kWh m$^{-2}$) (point out to daily radiation in the last three crop growing stages), and maximum, average and minimum temperatures (°C) which are recorded in the last three crop growing stages. It is aimed that finally to predict yield measured in (t ha$^{-1}$).

## III. HYBRID MACHINE LEARNING METHODS

As it is mentioned above, in this study, the performance of two hybrid machine learning methods in the crop yield prediction is compared. The selection of hybrid methods allows us to optimize the performance of machine learning in prediction [7]. According to Drummond et al. [8], ANN methods have had a very good performance in the crop yield prediction. Of course, the performance of neural networks can be determined by factors like the quality of the sample, the network structure, and the training parameters [9]. Therefore, in this study, hybrid methods of artificial neural networks-imperialist competitive algorithm (ANN-ICA) and artificial neural networks- gray wolf optimization (ANN-GWO) are applied. Following these methods are explained.

### A. Artificial Neural Network (ANN)

The ANN is a machine learning method that explores the relationship between phenomena (input-output data pairs) by benchmarking the human brain process in the problem-solving processes [9]. There are three layers in the ANN method input, hidden and output in which the connection among the layers provides the possibility of connecting every single neuron in one layer to all possible neurons in the other layer. The selections of and the design of meta-parameters such as learning rate, output function, number of nodes in the hidden layer, and inputs are very determinant in the accuracy and performance of the ANN [10].

In this method, the dataset is split into three categories, training, validation, and testing. The ANN learns interaction among input and out pairs by finding the patterns using algorithms such as back-propagation and optimization models [9], in the training phase. It is necessary to validate the datasets for adjusting and increase the accuracy of the learning process. And the prediction power of the developed ANN model is evaluated using the testing dataset and after the training phase.

### B. Imperialist Competitive Algorithm

The imperialist competitive algorithm is a well-known optimization algorithm [11] proposed by Atashpai Gargari and Lucas [12]. Atashpai Gargari and Lucas [12] inspired by imperialistic competition develop a method so-called imperialist competitive algorithm [13] in which the countries classified into two groups based on their power: 1) colonies and 2) imperialists. Where one empire includes an imperialist with its colonies. The empires tend to widen their territories by controlling more colonies from the other empires. This makes competition among the empires in which the strongest empire dominates and controls the weaker colonies [14].

### C. Hybrid ANN-ICA Method

Through hybridization of ANNs the parameters of the model can be optimized using an efficient optimization method. Here, the ICA is used to tune ANN meta-parameters optimally to improve the model accuracy [15]. In other words, ICA helps the ANN to optimize the weights and biases. As a result, the error rate decreases, and the performance of the model in prediction increases. In the literature, eight steps are introduced to perform an ICA: 1) defining an initial empire, 2) determining the positions of the imperialist and colonies, 3) specifying the power of an empire, 4) identifying the competition among the empires, 5) omitting the weaker empires, 6) the convergence [16]. Thus, if the meta-parameters (countries) are coded as the variables, then country X is defined as:

$$X \in \{x_1, x_2, x_3, x_4\} \quad (1)$$

The number of neurons in the hidden layer is shown as $x_1$. The $x_1$, can take a value of $1 < x_1 < 100$, $x_2$ is the input layer, and $x_3$ is the output activation function. The value of $x_1$ and $x_2$ can vary from 1 to 5 and $x_4$ is the learning rate that can be between $0 - 5$.

### D. Gray Wolf Optimizer

Mirjalili et al. [17] have developed the Gray Wolf Optimizer (GWO) algorithm inspired by the hunting process of gray wolves. Wolves usually lives in a group of 5 to 12 as two of them lead the group as it has formed a strong social hierarchy within the group. Mirjalili et al. [17] explain this hierarchy in a way that the alphas wolves (α) are the leader's group and make the decisions. The betas wolves (β) are in the second level and they support the alphas wolves' decisions. The deltas wolves (δ) are in the next level and they are the followers of alpha and beta wolves. According to Mirjalili et al. [17], there are 5 types of deltas wolves: 1) Scouts who are responsible for controlling the boundaries of the territory, 2) Sentinels who are the group security and protect the group in case of danger, Elders who are young and very strong and they are the potential alpha or beta wolves, Hunters who assist the alpha and beta in hunting prey for the group, and Caretakers who look after the ill and wounded wolves. In the lowest level of this community are the omegas wolves (ω) who follow the superior wolves and they are the lasts allowing to eat. Accordingly, alpha (α) wolves are the most desirable solutions in the GWO algorithm and the other best solutions are considered Beta (β) and delta (δ). In this methodology, the provided results in the group (populations) are omega (ω).

Following the mathematical equations of the hunting, process is provided.

Alpha, beta and omega wolves respectively have the responsibility to guide hunting. They firstly circle the pray's first step of hunting prey is circling it by α, β, and ω. The mathematical model of the circling process as shown in equations 1, 2, 3 and 4.

$$X(t+1) = X_p(t) + A.D \qquad (2)$$

Where X is the representation of the gray wolf position, t is the number of iterations, Xp is the prey position and finally, A and D can be calculated by (3) and (4).

In equation 4, represents the number of iterations, which varies from 0 to 2. *NumIter* refers to the total number of iterations. In addition, r1 and r2 are random vectors between [0,1] simulating the hunting. Equation 7 shows the update of wolves' positions.

$$D = |C.X_p(t+1) - X(t)| \qquad (3)$$

$$A = 2a.r_1 - a \qquad (4)$$

$$C = 2r_2 \qquad (5)$$

$$a = 2 - t(2/NumIter) \qquad (6)$$

$$X(t+1) = (X_1 + X_2 + X_3)/3 \qquad (7)$$

$X_1$, $X_2$, and $X_3$ can be calculated as follow:

$$X_1 = [X_\alpha - A_1.D_\alpha] \qquad (8)$$

$$X_2 = [X_\beta - A_2.D_\beta] \qquad (9)$$

$$X_3 = [X_\delta - A_3.D_\delta] \qquad (10)$$

$$D_\alpha = [C_1.X_\alpha - X] \qquad (11)$$

$$D_\beta = [C_2.X_\beta - X] \qquad (12)$$

$$D_\delta = [C_3.X_\delta - X] \qquad (13)$$

It can be interpreted that $X_1$, $X_2$, and $X_3$ are considered as the best solutions at t iteration. $A_1$, $A_1$, and A3 can be calculated by equation 3. And $C_1$, $C_2$, and $C_3$ are measured by equation 4.

*E. Gray Wolf Optimizer of Neural Networks*

In the gray wolf optimizer of neural networks (ANN-GWO) firstly, GWO trains the ANN to optimize the initial weight and biases. Then the neural network will be trained by the back-propagation algorithm to tune the weights and biases calculated in the previous stage in order to find the most global optima model.

*F. Accuracy metrics*

The next step modeling the machine learning methods is to test the accuracy. The model with the lowest error level and highest correlation will be selected as the best model. To assess the accuracy performance of the models of ANN-ICA and ANN-GWO, two metrics of Root mean square error (RMSE) and the relative mean absolute error [12] are selected. To measure the correlation, the metrics of correlation coefficient (R) is used. Equations 14, 15, and 16 explained these metrics in detail.

$$RMSE = \sqrt{\frac{1}{N}\sum_{i=1}^{N}(A-P)^2} \qquad (14)$$

$$R = \left(1 - \left(\frac{\sum_{i=1}^{n}(A-P)^2}{\sum_{i=1}^{n}A_i^2}\right)\right)^{1/2} \qquad (15)$$

$$MAE = \frac{\sum_{i=1}^{n}|A-P|}{N} \qquad (16)$$

Where A is the target value, P is the predicted values, and N is the numbers of data.

## IV. RESULTS AND DISCUSSION

As it is mentioned above, RMSE, MEA, and R metrics applied to test the accuracy of the ANN-ICA and ANN-GWO in the crop-yield prediction. As it is presented at the bottom of table II, the average RMSE of ANN-ICA (3.20) which is slightly higher than ANN-GWO (3.19). These figures indicate that, in terms of these two metrics, the performance of ANN-GWO in the crop yield production has been better than ANN-ICA. The best result per crop is highlighted in bold in table II.

Besides, the results for the metric of R depicted in Fig. 2. Here best result per crop is highlighted in bold in the table as well. Since the average metric of R is higher for ANN-GWO (0.48) method than ANN-ICA (0.42) it is interpreted that the performance of ANN-GWO in this metric has been better. In addition, the average MEA for ANN-ICA (27.22) is higher than ANN-GWO (26.65) which represents the better performance of ANN-GWO in these metrics. According to table II and the R metric, an individual counting of best results reveals that ANN-GWA gets most of the best-correlated models (three models).

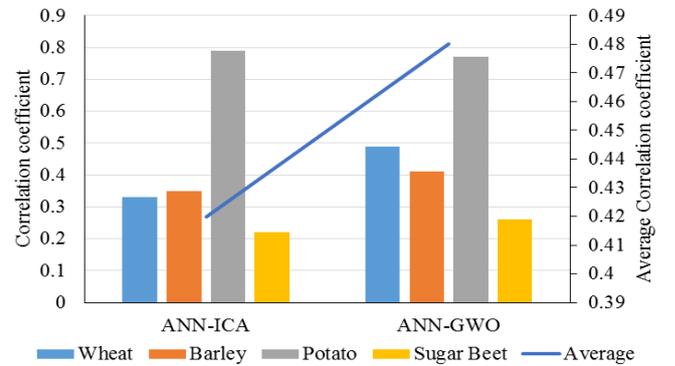

Fig. 2. Comparison of correlation coefficient of ANN-ICA and ANN-GWO per each crop.

Since selected attributes affect the model performance [18] the importance of each attribute in the performance of the selected methods (ANN-ICA and ANN-GWO) is measured and presented in table III.

To simplify presenting the attributes in the table, each attribute gets a code. Attribute code for attributes planting area is AT1, for irrigation water depth is AT2, for rainfall during the crop growth stages is AT3, for global solar radiation is AT4, for maximum temperatures is AT5, for average temperatures is AT6, and for minimum temperatures is AT7. One of our contributions to the current research is that the attributes are compared, and the best attribute set is identified for each technique. Table III exposes the lack of consistency among the best attribute sets chosen by the methods as there is no evidence of proving the preference of an individual model to constitute a set of attributes. Whilst, there is in the set of attributes by the methods related to the crops. For example, for all the crops AT2 are included regardless of the methods.

TABLE II. EVALUATION OF THE ACCURACY OF ANN-ICA AND ANN-GWO WITH R, MEA, AND RMSE

| Crops | R | | MEA (%) | | RMSE (%) | |
|---|---|---|---|---|---|---|
| | ANN-ICA | ANN-GWO | ANN-ICA | ANN-GWO | ANN-ICA | ANN-GWO |
| Wheat | 0.33 | **0.49** | 35.04 | **33.30** | 8.48 | **8.41** |
| Barley | 0.35 | **0.41** | **12.07** | 12.13 | **0.32** | 0.33 |
| Potato | **0.79** | 0.77 | 22.76 | **22.15** | 0.68 | **0.66** |
| Sugar Beet | 0.22 | **0.26** | **38.99** | 39.02 | **3.32** | 3.34 |
| Average | 0.42 | 0.48 | 27.22 | 26.65 | 3.20 | 3.19 |

TABLE III. THE EFFECT OF ATTRIBUTES ON THE ANN-ICA AND ANN-WGO MODELS

| Crops | Method | Attributes | | | | | | |
|---|---|---|---|---|---|---|---|---|
| | | AT1 | AT2 | AT3 | AT4 | AT5 | AT6 | AT7 |
| Wheat | ANN-ICA | 0 | 1 | **3** | 1 | 0 | 0 | 1 |
| | ANN-GWO | 0 | **3** | 0 | 2 | 2 | 0 | 1 |
| Barley | ANN-ICA | 1 | **3** | 1 | 2 | 1 | 1 | 0 |
| | ANN-GWO | 0 | 2 | 1 | 1 | **3** | 1 | 0 |
| Potato | ANN-ICA | **3** | **3** | **3** | 0 | **3** | **3** | 2 |
| | ANN-GWO | **3** | **3** | **3** | 0 | 1 | 1 | 2 |
| Sugar Beet | ANN-ICA | 2 | **3** | 2 | 1 | 0 | 1 | 2 |
| | ANN-GWO | 2 | **3** | 0 | 2 | 1 | **3** | 0 |

## V. CONCLUSION

Various approaches are taken by researchers to predict the crop yield, such as regression models and machine learning methods. The main contribution of the current study and comparison of two hybrid machine learning methods for the prediction of crop yield, for the first time in the literature. One of the most important reasons that hybrid machine learning methods are considered is that the accuracy of the prediction of such models is higher. On the other hand, in this study, numerous attributes are considered to evaluate the performance of the models. For this purpose, firstly, the best attributes set for each method were identified among the potential attributes. According to the result the ANN-GWO method with R of 0.48, RMSE of 3.19, and MEA of 26.65 had a better performance in the crop yield prediction. Since, a different set of attributes affect the performance of the model, it is recommended that future research examine a different set of attributes and compare the results. Besides, it is also recommended to compare other hybrid machine learnings to find the proper model. For the future research, advancement of hybrid and ensemble machine learning models, e.g., [19-24], and comparative analysis with deep learning models, e.g., [25-28] are proposed to identify models with higher efficiency.

ACKNOWLEDGMENT

We acknowledge the financial support of this work by the Hungarian State and the European Union under the EFOP-3.6.1-16-2016-00010 project and the 2017-1.3.1-VKE-2017-00025 project.

.